# ParamBench: A Graduate-Level Benchmark for Evaluating LLM Understanding on Indic Subjects

**Ayush Maheshwari, Kaushal Sharma[♠], Vivek Patel[♠], Aditya Maheshwari[♠]**
[♠]Indian Institute of Management Indore, India ; [♠]BharatGen
ayush.hakmn@gmail.com, {kaushals,vivekp,adityam}@iimidr.ac.in

## Abstract

Large language models have been widely evaluated on tasks such as comprehension, summarization, code generation, *etc*. However, their performance on graduate-level, culturally grounded questions in the Indian context remains largely unexplored. Existing Indian benchmarks emphasise basic fact-orientated queries that offer limited assessment of a deeper disciplinary understanding tailored to the Indian setting. In this paper, we present ParamBench, consisting of more than 17K questions in the Hindi language, comprising questionnaires from 21 diverse subjects. These questions are primarily derived from a nationwide graduate-level entrance examination covering topics such as history, music, instruments, yoga, literature, philosophy, law, *etc*. specifically for the Indian context. Additionally, we assess the ability of LLMs to handle diverse question formats—such as list-based matching, assertion–reason pairs, and sequence ordering—alongside conventional multiple-choice questions. We evaluated the performance of more than 16 open source LLMs on this benchmark, observing that Gemma3-27B attains the highest overall accuracy of 56.4%. Furthermore, subject-wise analysis indicates that even for the best-performing LLMs, performance remains weak on topics such as music, classical instruments, and law, underscoring persistent challenges in culturally grounded reasoning. The dataset and source code is present at https://github.com/ayushbits/ParamBench.

## 1 Introduction

Large Language Models (LLMs) have demonstrated remarkable capabilities in multilingual reasoning and knowledge-intensive tasks (Liu et al., 2024). Although LLMs perform reasonably well in English and a few other languages, their performance in culturally nuanced domains, particularly within the Indian context, remains weak (Verma et al., 2025). This is especially significant given India's linguistic and cultural diversity, with a population of over 1.4 billion, more than 120 major languages, and nearly 19,500 dialects across 28 states (Javed et al., 2024). Without robust evaluation in these settings, the application of LLMs to education, governance, and knowledge systems in India risks being incomplete and inequitable.

India has a rich body of traditional knowledge in several areas such as history, religion, law, literature, philosophy, music, medicine, *etc*. Yet state-of-the-art LLMs often perform poorly when questions are related to familiarity with indigenous conceptual frameworks and knowledge (Maji et al., 2025). These weaknesses become clear in tasks that require an understanding of Indian ways of thinking, local concepts, and culturally specific knowledge. Existing Indic language benchmarks, while valuable for assessing syntactic and task-oriented competencies (Doddapaneni et al., 2023; Verma et al., 2025), fail to capture the diverse nuances. Recent resources such as the Sanskriti dataset (Maji et al., 2025) capture culturally salient attributes across India's geographic diversity. However, their emphasis on breadth leaves a gap for evaluation on in-depth graduate-level knowledge in culturally aligned subjects.

To address this gap, we present a new benchmark in the Hindi language, ParamBench consisting of more than 17K questions across 21 India-focused subject areas—including archaeology, religion, law, culture, music, arts, philosophy, and yoga, *etc*. (*c.f.* Figure 2). The questions are drawn from postgraduate-level competitive exams and reflect fields grounded in India's intellectual and cultural traditions. Additionally, the benchmark includes multiple question types: standard multiple-choice, list-based matching, assertion–reason, sequencing/ordering, incorrect-statement identification, and fill-in-the-blank. The benchmark tasks examine not only language understanding in In-dian languages but also whether models can understand and use concepts that are specific to Indian history, music, philosophy, law, literature, arts, *etc*. We provide example instances from ParamBenchfor different questions types in Table 8 in Appendix.

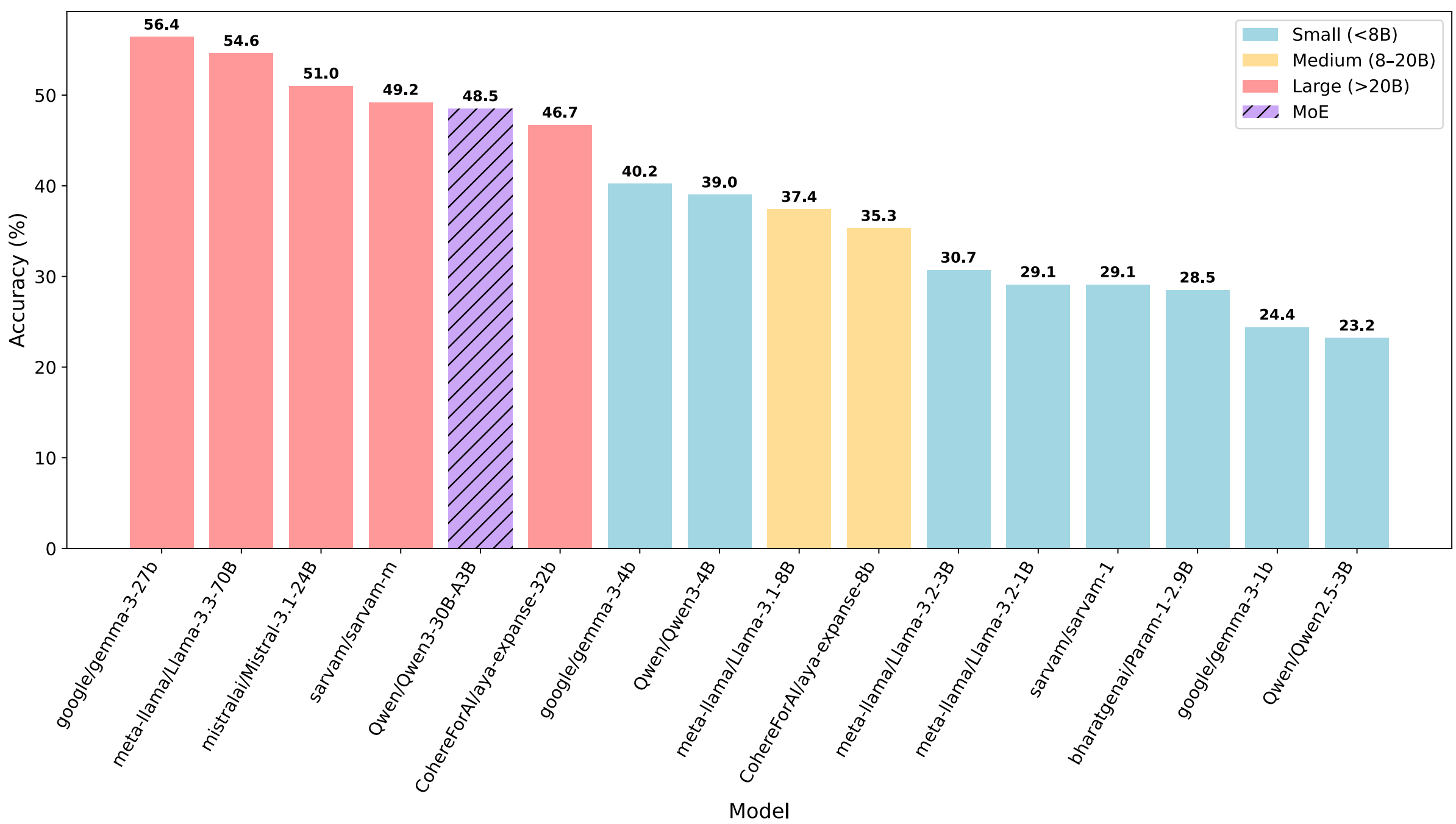

Figure 1: Average performance of evaluated models on ParamBench, categorized by the parameter sizes.

In ParamBench, we evaluate around 16 LLMs of different model families and parameter sizes, including recently released Indian LLMs. While many contemporary models achieve strong scores on standard English-centric benchmarks—covering general reasoning, question answering, and reading comprehension—their accuracy declines substantially on culturally grounded Indic topics (*c.f.* Figure 1). Among smaller models with fewer than 8 billion parameters, Gemma3-4B achieves the best overall accuracy at 40.2%. In the category of larger models, Gemma3-27B attains the best accuracy of 56.4%. Notably, the MoE model Qwen3-30B-A3B substantially outperforms the models with <8B params, indicating that expert routing captures a broader range of knowledge than comparably dense models despite activating only a fraction of parameters per token. However, performance on several Indic subjects—including music, Indian culture, drama, archaeology, law, and traditional instruments—remains below 52% even for the best-performing model (*c.f.* Table 2). This highlights the persistent challenges in culturally nuanced domains despite increased scale and instruction tuning. Our contributions can be summarised as follows:

1. We present ParamBench, consisting of around 17.2K questions spanning 16 subjects related to indian knowledge systems such as yoga, music and instruments, law, drama and theatre, Indian culture, archaeology, *etc*.

2. We systematically evaluate several open LLMs on different question types that include MCQs, list matching, assertion and reasoning, ordering and incorrect statement identification.

3. We examine how performance scales with model size, observing consistent and significant gains as the number of parameters increases. In addition, we conduct a systematic analysis of model behaviour across subject areas, identifying both relative strengths in certain domains and weaknesses in culturally grounded subjects.

With ParamBench, our aim is to identify and quantify current gaps in LLM performance for the Indian context and guide the development of models that are culturally and linguistically aligned with India. Our goal is to help build AI systems that better represent India's knowledge traditions and language diversity.

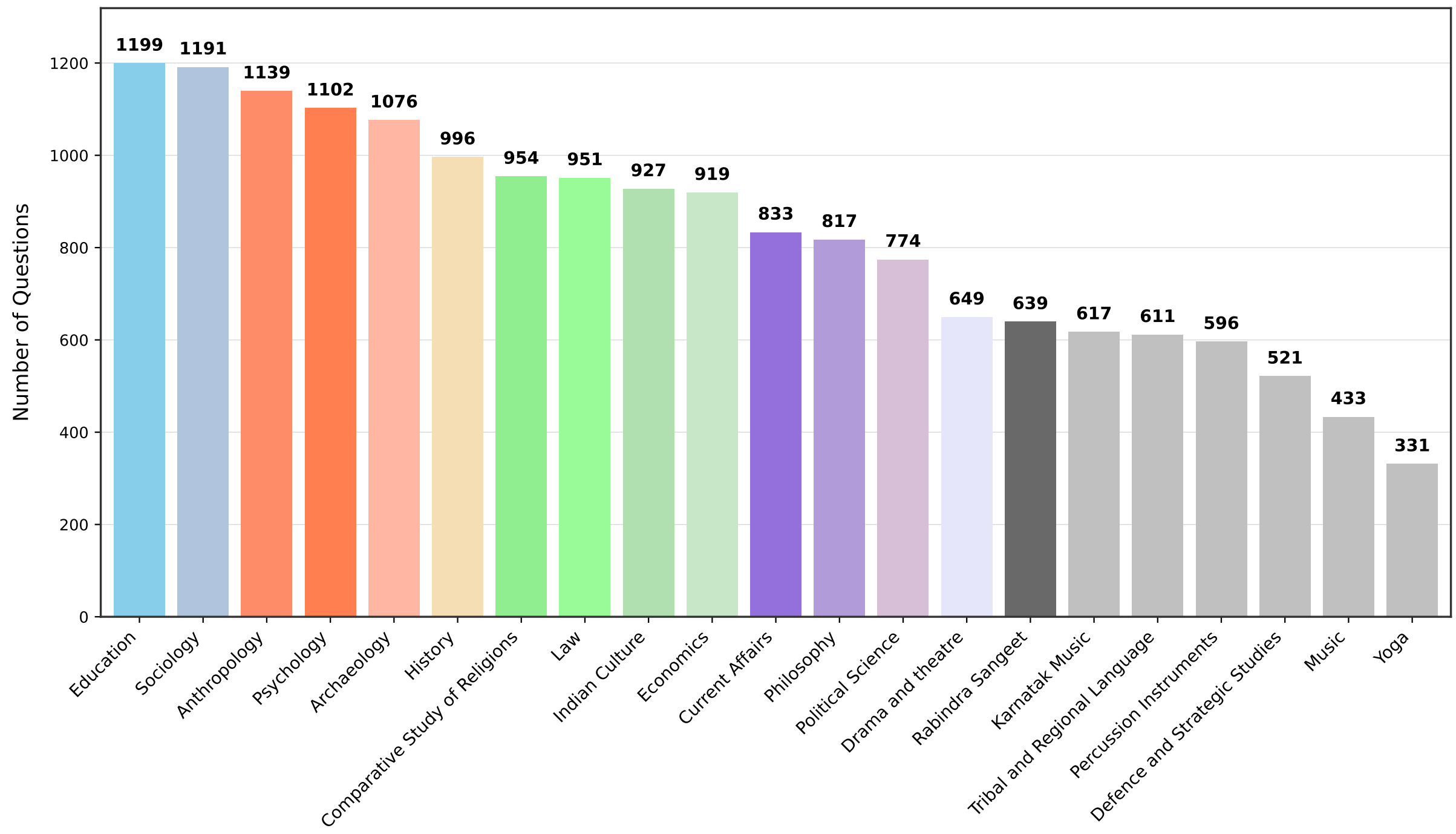

Figure 2: Distribution of the questions across different subjects.

## 2 Related Works

### 2.1 LLM Benchmarks

Several benchmarks have been developed to evaluate the capabilities of large language models (LLMs). For instance, KMMLU and CMMLU (Son et al., 2025; Li et al., 2024) assess academic knowledge across a broad range of subjects, while BIG-Bench (Srivastava et al., 2023) focuses on measuring complex reasoning and generalization abilities. Similarly, HELM (Liang et al., 2023) introduces a comprehensive framework that evaluates LLMs across multiple dimensions, including accuracy, robustness, and fairness.

Although these benchmarks provide extensive coverage of topics, they remain largely centered on English and other high-resource languages. Consequently, they often overlook linguistic and cultural diversity. Recent LLMs demonstrate some capacity for cultural and linguistic knowledge (Johnson et al., 2022; Atari et al., 2023; Masoud et al., 2025), yet continue to face significant challenges in adapting to non-Western contexts (Alkhamissi et al., 2024; Durmus et al., 2024).

### 2.2 Indian multilingual benchmarks

In recent years, various benchmarks have been introduced to evaluate LLMs in the context of Indian languages and multilingual tasks. The IndicQA Benchmark (Singh et al., 2025) provides large-scale question-answering datasets in 11 Indic languages, incorporating both original and translated content. MILU (Verma et al., 2025) expands this effort by presenting more than 80,000 multiple-choice questions in 11 languages, with a particular emphasis on culturally relevant topics. Similarly, IndicGenBench (Singh et al., 2024) focuses on generative tasks, such as summarization and translation, spanning 29 Indic languages. BharatBench (Krutrim, 2025) broadens the scope further by integrating text, vision, and speech modalities across 8 Indian languages.

More recently, cultural and evaluator alignment have been highlighted through benchmarks such as SANSKRITI (Maji et al., 2025) and PARIKSHA (Watts et al., 2024), which address the need to incorporate socio-cultural context in evaluating LLMs. Beyond task-specific benchmarks, several benchmarks have been proposed for extremely low-resource Indic languages such as Sanskrit (Maheshwari et al., 2022, 2024).

In science and technical education, JEEBench (Arora et al., 2023) evaluates engineering entrance-level mathematics, and the Materials Science Graduate Exam Benchmark targets post-graduate level scientific knowledge. In the legal and finance domain, IL-TUR (Joshi et al., 2024) assesses legal reasoning, while LLMs

Acing Chartered Accountancy (Gupta et al., 2025) evaluates performance on taxation and auditing tasks. For governance and multilingual reasoning, datasets such as the UPSC Civil Services Study dataset (Banerjee et al., 2024), MILU (Verma et al., 2025), and IndicMMLU-Pro (KJ et al., 2025) examine general knowledge and reasoning across multiple Indic languages. Building on these efforts, our benchmark contributes by specifically evaluating India-centric knowledge through expert-verified multiple-choice questions (MCQs) drawn from the UGC-NET and UPSC examinations in Hindi.

## 3 ParamBench

In this section, we present data collection, annotation process and analysis.

### 3.1 Data Collection

ParamBench consists of 17,275 questions in the is Hindi language covering 21 Indic subjects such as Indian history, literature, archaeology, Indian culture, music, arts, yoga, *etc*. The subject-wise distribution of questions are present in Figure 2. The questions are collected from UGC-NET[1] and UPSC Civil services examination[2]. UGC-NET is a nationwide examination administered by a government agency to determine eligibility for PhD admission and for appointment to teaching positions in Indian universities and colleges. The exam is offered in around 85 subjects and is conducted twice annually. Each test consists of two papers composed of multiple-choice questions (MCQs). UPSC Civil Services likewise employs rigorous multiple-choice assessments as part of a multi-stage selection process, providing domain-relevant, exam-realistic materials for evaluating graduate-level competence in India-specific subjects.

We constructed the dataset by downloading official question papers and answer keys from the respective examination websites. For UGC-NET, we curate papers from 2012–2018, selecting questions from 21 subjects that relate to Indian knowledge, including Indian history, law, music, and culture. We did not include other subjects as those are partially covered by other existing benchmarks such as Sanskriti (Maji et al., 2025), MILU (Verma et al., 2025).

[1] https://ugcnet.nta.ac.in

[2] https://upsc.gov.in/examinations

For UPSC, we include preliminary examination papers from 2011–2024, focusing on six major subjects that are central to Indian civilizational, literary, cultural, and academic knowledge. Each subject comprises multiple question papers in PDF, in which many of them are machine-readable, while a subset contains non-selectable text. Layouts vary across documents, with some in single-column format and the majority in two-column format. To ensure uniform text accessibility, we processed all PDFs with a proprietary OCR system and obtained text outputs for downstream curation and annotation.

To the best of our knowledge, this corpus has not appeared in prior LLM benchmarking studies and constitutes a newly curated, human-authored dataset designed explicitly for graduate-level evaluation in Indic contexts.

### 3.2 Annotation setup

We process the OCR and extract text by tagging each question with its subject and the exam year of appearance. Following this, human annotators perform post-OCR correction to fix recognition errors and restore missing diacritics or script artefacts. Beyond textual corrections, annotators standardise formatting so that questions and answers are parseable by automated scripts. This pipeline ensures consistent metadata, clean text, and machine-actionable structure across heterogeneous source documents. Each entry was structured with fields for question, question type, options, correct answer, subject, year, and exam name to ensure consistency and traceability.

Annotation was conducted by subject-matter experts proficient in Hindi and trained in the relevant domains. Because the source questions and answers are in Hindi, annotators were selected based on demonstrated fluency in reading, writing and speaking of Hindi and grammar knowledge.

General annotation guidelines were developed and shared with all annotators. These emphasised grammatical correctness, completeness of questions and answers, and standard formatting. Questions with unresolved issues were to be removed or escalated for review. Annotation was primarily carried out using Google Docs, and the finalized dataset was exported in CSV format.

### 3.3 Team structure

The annotation team consisted of two tiers. First, a team of four annotators corrected questions and answers from OCR outputs, entered answer keys, and corrected grammatical errors. This was followed by a review process by a subject expert, who verified the grammar, formatting, and correctness of the answer keys.

We implemented both manual and automated quality assurance protocols. Initially, only those questions that aligned with the benchmark's focus on Indian-specific knowledge were retained. Manual checks were performed to validate grammatical correctness, answer accuracy, and completeness. Automated scripts were used to ensure each question had exactly four options, one correct answer, and no missing fields. This two-step quality control process helped maintain the reliability of the annotated dataset.

Prior to full-scale annotation, annotators were given a sample dataset along with worked examples. They were then assigned trial files, which were reviewed and corrected by the reviewers. Feedback was provided iteratively, and only after demonstrating consistent accuracy were annotators assigned larger batches of data. All annotation work was performed using Google Spreadsheets. Annotators were compensated at a rate of $1 per 10 questions.

| Question Type | # Questions |
|---|---|
| MCQ | 10668 |
| Match the List | 2227 |
| Assertion and Reason | 1855 |
| Find incorrect Statement | 1407 |
| Ordering | 1072 |
| Fill in the Blank | 46 |
| **Total** | **17275** |

Table 1: Distribution of question types in ParamBench

### 3.4 Statistics

Figure 2 presents the overall distribution of questions in ParamBench. The corpus comprises 17,275 Hindi questions spanning 21 subjects. Education contributes the largest share (1199 questions), whereas Yoga has the fewest (330 questions). The median and mean number of questions per subject are 833 and 822.6, respectively. This also underlines an almost symmetric spread of the number of questions across the subject domains. The standard deviation is around 254.26, and the coefficient of variation is 30.9%, which tells that the questions are not evenly spread across the domain, but there is an absence of highly significant variance. It can also be noted that the skewness of the question length is pretty high, with 181.7 mean characters per question compared with a median of only 95 characters per question. On the other hand, the option length is close to symmetric, with a mean of 21.2 characters per option and a median of 15.5 characters per option.

Notably, the benchmark includes substantial coverage of Indic domains that are rarely represented in prior evaluations, such as drama and theatre, Rabindra Sangeet, Tribal and Regional Literature, Percussion Instruments, and Yoga. Although some categories exhibit thematic overlap—for example, Music with Carnatic music and Rabindra Sangeet—we retain them as distinct subjects to preserve domain specificity and enable fine-grained analysis.

We categorize each item in the table by question type to reflect the range of exam-style reasoning (see Table 1), the majority of which are multiple-choice questions (10,668). The remaining items comprise Match-the-List (2,227), Assertion–Reason (1,855), Find-the-Incorrect-Statement (1,407), Sequencing/Ordering (1,072), and a small Fill-in-the-Blank subset (46). This composition emphasizes selection-based reasoning while providing complementary coverage of mapping/alignment (list matching), causal justification (assertion–reason), error detection (incorrect statement), temporal ordering (sequencing), and limited cloze-style completion (fill-in-the-blank).

## 4 Experiments

We evaluate ParamBench on 16 openly available models spanning diverse sizes and architectures, including parameter ranges of 1B–8B, 8B–15B, and over 15B, as well as a Mixture-of-Experts (MoE) model. All experiments follow a zero-shot setup using the prompt in Table 5, without demonstrations or validation examples. Models are retrieved from publicly available checkpoints at HuggingFace and executed by using the transformers library. We use greedy decoding while generating predictions by setting temperature as 0 and do_sample flag to false. During inference, we set the batch size as 16. We disable thinking mode for Qwen3-MoE. For Gemma and Mistral, we follow their respective strategies from response generation instead of pipeline function.

Evaluation is based on direct answer matching generated responses are compared with the groundtruth answer, and accuracy is reported as the proportion of correct outputs. This yields a clear and interpretable measure of task performance in realistic settings.

All models are instruction-tuned variants. We also performed experiments with pre-trained base variants of these models, but performance was expectedly worse than instruction-tuned variants. The following models were used in our evaluation:

1. **Llama Series:** We have used the Llama 3 collection, which offers several multilingual language models (Team, 2024). We used Llama-3.2-1B, Llama-3.2-3B, Llama-3.1-8B, Llama3.3-70B during experiments. These models were trained on approximately 15 trillion tokens.

2. **Qwen series:** We evaluated Qwen2.5-3B and Qwen3-4B-Instruct-2507. The latter represents a substantial improvement over Qwen3-4B and operates exclusively in non-thinking mode. Additionally, we conducted evaluations on the MoE Qwen3-30B-A3B model, which comprises a total of 30 billion parameters with 3 billion active parameters. The pretraining process for Qwen3 uses a large-scale dataset of approximately 36 trillion tokens, covering 119 languages and dialects (Team-Qwen3, 2025).

3. **Gemma Series:** We evaluated several models from the Gemma series, which have demonstrated strong performance on Indic languages. Specifically, we employed the Gemma-3 series models with 1B, 4B, and 27B parameters. The 27B model was pretrained on 14T tokens, while 4B and 1B parameter models were pre-trained on 4T and 2T tokens, respectively (Team-Gemma2, 2024).

4. **Mistral 3.1** - We evaluate over mistral-small-3.1-24B-Instruct-2503 instruction-tuned model containing 24B parameters[3].

5. **Cohere series:** We evaluated over aya-expanse 8B and 32B parameter models, which have demonstrated strong multilingual capabilities. (Aryabumi et al., 2024).

6. **Sarvam Models:** Sarvam-1 is 2B parameter language models specifically optimized for Indian languages. We also evaluate Sarvam-M, which is a 24B parameter model post-trained on Mistral-Small-3.1. Although multiple fine-tuned Indic LLM variants are available, we selected this model because it demonstrates the highest performance among them on Indic benchmarks (SarvamAI).

7. **Param-1-2.9B:** PARAM-1 a bilingual language model trained from scratch in English and Hindi containing 2.9 billion parameters. The model is trained on 7.5 Trillion tokens in English and Hindi languages (Pundalik et al., 2025).

[3] https://mistral.ai/news/mistral-small-3-1

## 5 Result

### 5.1 Overall Model Performance

The evaluation results in Figure 1 show that the overall accuracies remain modest across various model sizes. In the category of models with fewer than 8B parameters, the highest performance was observed with gemma3-4b (40.2%) (Team, 2024), while most others, including Llama3.2-3B and Sarvam-1, stayed around 30%. This suggests that even with small-model capacity, language and data mixture play a significant role in capturing the breadth of culturally grounded knowledge. Similarly, among mid-sized models (8B–15B parameters), average performance is improved for both Llama-3.1-8B (37.4%) and aya-8 B (35.3%); however, it cannot surpass the best-performing Gemma-4 B and Qwen3-4 B models. This points to the importance of pretraining mixture and domain coverage over parameter count.

Among large models (>20B), Gemma3-27B tops the chart at 56.4%, outscoring the 70B Llama variant in the pool, which underscores that scale without targeted language/domain exposure may underperform a smaller model whose pretraining better aligns with Indic distributions. The MoE model Qwen3-30B-A3B attains 48.5% with only 3B active parameters per token, highlighting that sparse expert routing can yield competitive accuracy on Indic topics at substantially lower activated capacity.

### 5.2 Subject-wise Analysis

In Table 2, we provide subject-wise analysis for the larger size LLMs. Current Affairs demonstrates the strongest performance overall, with Llama-3.3-70B achieving 90.2% accuracy, closely followed by Gemma-3-27B at 90%. This suggests that models perform well on topics with broad coverage in training datasets and contemporary relevance. Interestingly, Gemma-3-4B achieve 70.7% in Current Affairs despite its smaller size. The questions in Defense and Strategic Studies overlap significantly with current affairs, involving contemporary military developments, strategic policies, and geopolitical events that receive substantial coverage in news media and policy discussions. This explains why models perform relatively well in this domain compared to more specialized subjects. Education also shows relatively strong performance across models, with Gemma-3-27B reaching 65.1% and Llama-3.3-70B at 62.1%, suggesting that pedagogical content receives reasonable representation in pretraining corpora.

| Subject | Aya-8b | Llama-8B | Mistral3.1 | Sarvam-m | Gemma27b | Qwen-MoE | Aya-32B | Llama-70B |
|---|---|---|---|---|---|---|---|---|
| Current Affairs | 63 | 73.6 | 88.5 | 88.2 | 90 | 66.1 | 78.4 | **90.2** |
| Comparative Religion | 45.6 | 49.1 | 65.5 | 60.5 | **70.3** | 64.5 | 64.6 | 68.9 |
| Defence Studies | 42.2 | 43.6 | 58.3 | 59.1 | **70.2** | 60.7 | 60.3 | 67.4 |
| Education | 41.5 | 43.6 | 55.9 | 56.5 | **65.1** | 59.4 | 52.5 | 62.1 |
| Sociology | 39 | 37.4 | 54.1 | 55 | 63.1 | 52.6 | 51.2 | **63.1** |
| Yoga | 38.7 | 38.1 | 57.4 | 56.2 | **63.1** | 52.6 | 49.8 | 62.8 |
| Anthropology | 32 | 34.6 | 52.4 | 50.7 | **60** | 49.5 | 44.7 | 56.9 |
| Tribal Language | 38.3 | 43.5 | 56 | 52.7 | 59.1 | 54.5 | 53.7 | **59.1** |
| Psychology | 33.2 | 31.7 | 47.9 | 47.2 | **57.5** | 50.2 | 43.8 | 53.6 |
| Economics | 36.5 | 36 | 46.1 | 46.4 | 56.3 | 52.6 | 45.2 | **54.6** |
| Political Science | 31 | 32.4 | 49.5 | 47.9 | **55.2** | 45.9 | 43.9 | 52.3 |
| Philosophy | 33.5 | 36.4 | 49.9 | 45 | 53.5 | 47.2 | 45.9 | **53.7** |
| History | 29.5 | 34.3 | 52.1 | 45.8 | **52** | 42.2 | 42.9 | 49.9 |
| Indian Culture | 25.5 | 33 | 47.2 | 43.4 | **51.9** | 43.5 | 41.1 | 50.3 |
| Archaeology | 31.6 | 31.8 | 45.9 | 44.7 | **51.6** | 41.9 | 38 | 47.9 |
| Drama & theatre | 33.9 | 32.7 | 47.8 | 45.9 | **51.2** | 47.9 | 39.6 | 47.8 |
| Law | 32.6 | 33.4 | 44 | 41.9 | 48.2 | 41.9 | 41.1 | **48.5** |
| Rabindra Sangeet | 24.4 | 28.8 | 31.6 | 34.6 | **36.9** | 31.6 | 32.6 | 32.2 |
| Music | 26.1 | 25.4 | **36.3** | 33.5 | 36.3 | 30.3 | 35.6 | 35.8 |
| Percussion | 29.7 | 31 | 31.9 | 30.5 | **35.9** | 34.1 | 33.7 | 33.7 |
| Karnatak Music | 28.7 | 28.2 | 36 | 32.9 | 33.5 | 31.1 | 32.4 | **34.4** |

Table 2: We report the accuracy across different subjects for LLMs with parameter size >8B. The best-performing model for each subject is indicated in **bold**.

Music-related subjects remain challenging across all model sizes. Performance in Music ranges from 28.2% (Gemma-3-4B) to 36.9% (Gemma-3-27B), while Karnatak Music shows similar patterns with scores between 31.1% and 36%. Percussion Instruments follows the same trend, with accuracies spanning 31.1% to 35.9%. These consistently lower scores indicate that specialized musical terminology, cultural context, and domain-specific knowledge are substantially underrepresented in current pretraining mixtures.

We observe that Sarvam-M, a post-trained version of Mistral-Small-3.1-24B specifically optimized for Indic languages, achieves performance levels closely comparable to its base model across most evaluated categories. The results demonstrate that Sarvam-M maintains competitive accuracy with Mistral-3.1-24B on the majority of subjects-for instance, both models score identically on Current Affairs (88.2 vs 88.5) and show minimal differences on Defence Studies (59.1 vs 58.3) and Education (56.5 vs 55.9). This suggests that post-training over already stronger models having good coverage of Indic subjects might not be very beneficial for our ParamBench.

### 5.3 Question type-wise results

In Table 3, we present question-type analysis across the top-performing models. Since all evaluated models are instruction-tuned variants, the variation across question types likely reflects differences in supervised fine-tuning (SFT) strategies and the composition of instruction-following datasets rather than raw pretraining alone.

Assertion and Reason questions show the strongest performance overall, with Llama-3.3-70B leading at 61.8%, followed by Gemma-3-27B at 60.5%. Aya-32B achieves 54.4%, while the smaller models—Sarvam-M (51.6), Qwen3-MoE (47.2), and Gemma-3-4B (46.5). This format requires evaluating both a statement and its justification, which appears better aligned with models' reasoning capabilities compared to other question types.

| Question Type | Gemma-27b | Llama-70b | Mistral3.1 | Sarvam-m | Qwen3-MoE | Aya-32b | Gemma3-4B |
|---|---|---|---|---|---|---|---|
| Assertion & Reason | 60.5 | **61.8** | 44.9 | 51.6 | 47.2 | 54.4 | 46.5 |
| Incorrect Statement | **60.4** | 45.7 | 52.4 | 44.1 | 53.6 | 48.8 | 42.9 |
| Fill in the Blank | **58.7** | 54.3 | 50 | 45.7 | 54.3 | 43.5 | 52.2 |
| MCQ | 56.6 | **57.8** | 51.9 | 49.9 | 48.7 | 50.1 | 41.9 |
| List Matching | **52.7** | 41.9 | 50.7 | 48.8 | 45.5 | 29.3 | 28.2 |
| Ordering | **50.1** | 49 | 51 | 46.7 | 47.9 | 33.5 | 32.9 |

Table 3: We present the average accuracy across all subjects for various question types, focusing on the six top-performing models. The best-performing model for each question type is indicated in **bold**, while the worst-performing is underlined. Detailed results for each model are provided in Table 6 in the Appendix.

Identifying incorrect Statement questions reveals interesting disparities. Gemma-3-27B achieves the highest score at 60.4%, while Mistral-3.1-24B (52.4) and Qwen3-MoE (53.6) perform competitively. Notably, Llama-3.3-70B underperforms at 45.7% despite its size, trailing even Sarvam-M (44.1) and Aya-32B (48.8). This disparity suggests that error-detection capabilities depend heavily on SFT design.

Fill in the Blank questions show Gemma-3-27B leading at 58.7%, with Llama-3.3-70B and Qwen3-MoE as 54.3%. Mistral-3.1-24B achieves 50, while Sarvam-M (45.7) and Gemma-3-4B (52.2) demonstrate that smaller models can still perform reasonably on this cloze-style format with appropriate instruction tuning. Sequence/Ordering emerges as the most challenging format across all models. Gemma-3-27B leads at 50.1%, followed by Mistral-3.1-24B (51) and Llama-3.3-70B (49). Qwen3-MoE (47.9), Sarvam-M (46.7), Aya-32B (33.5), and Gemma-3-4B (32.9). This indicates that all LLMs struggle with temporal reasoning and procedural ordering regardless of architecture or scale.

Models handle assertion-logic and standard MCQs reasonably well, show variable capability on error detection and matching depending on SFT composition, but universally struggle with sequencing tasks—highlighting a gap in how current instruction datasets prepare models for structured temporal reasoning on Indic content

## 6 Conclusion

LLMs continue to struggle when evaluated on culturally grounded, India-specific domains despite strong performance on general benchmarks. ParamBench , containing >17K questions, fills this gap by offering a rigorous, graduate-level evaluation across 21 diverse subjects in the Hindi language rooted in India's intellectual traditions. Our results reveal clear performance drops across leading models, emphasizing the need for culturally aligned benchmarks. We envision ParamBench as both a diagnostic tool and a stepping stone toward developing LLMs that are more inclusive of India's linguistic and knowledge diversity.

## 7 Limitations

Our evaluation of ParamBench was conducted exclusively on openly available models, precluding assessment of proprietary systems such as OpenAI GPT-4, Anthropic Claude, or Google Gemini. Additionally, computational constraints limited our evaluation to models with fewer than 70B parameters. Consequently, we did not assess very-large models such as the recently released Qwen3-235B-A22B (235B parameters with 22B active) or Llama-3.1-405B (405B parameters), which may demonstrate different scaling behaviors on Indic content. Future work should investigate whether these frontier-scale models exhibit improved performance on culturally grounded benchmarks, as their massive capacity could potentially capture more comprehensive knowledge distributions across languages and domains.

## References

Badr Alkhamissi, Muhammad ElNokrashy, Mai Alkhamissi, and Mona Diab. 2024. Investigating cultural alignment of large language models. In *Proceedings of the 62nd Annual Meeting of the*

*Association for Computational Linguistics (Volume 1: Long Papers)*, pages 12404–12422.

Daman Arora, Himanshu Singh, et al. 2023. Have llms advanced enough? a challenging problem solving benchmark for large language models. In *Proceedings of the 2023 Conference on Empirical Methods in Natural Language Processing*, pages 7527–7543.

Viraat Aryabumi, John Dang, Dwarak Talupuru, Saurabh Dash, David Cairuz, Hangyu Lin, Bharat Venkitesh, Madeline Smith, Jon Ander Campos, Yi Chern Tan, Kelly Marchisio, Max Bartolo, Sebastian Ruder, Acyr Locatelli, Julia Kreutzer, Nick Frosst, Aidan Gomez, Phil Blunsom, Marzieh Fadaee, Ahmet Üstün, and Sara Hooker. 2024. Aya 23: Open weight releases to further multilingual progress.

Mohammad Atari, Jonathan Haidt, Jesse Graham, Sena Koleva, Sean T. Stevens, and Morteza Dehghani. 2023. Morality beyond the weird: How the nomological network of morality varies across cultures. *Journal of Personality and Social Psychology*, 125(5):1157–1188.

Somonnoy Banerjee, Sujan Dutta, Soumyajit Datta, and Ashiqur R KhudaBukhsh. 2024. Gender representation and bias in indian civil service mock interviews. *CoRR*.

Sumanth Doddapaneni, Rahul Aralikatte, Gowtham Ramesh, Shreya Goyal, Mitesh M Khapra, Anoop Kunchukuttan, and Pratyush Kumar. 2023. Towards leaving no indic language behind: Building monolingual corpora, benchmark and models for indic languages. In *Proceedings of the 61st Annual Meeting of the Association for Computational Linguistics (Volume 1: Long Papers)*, pages 12402–12426.

Esin Durmus, Karina Nguyen, Thomas Liao, Nicholas Schiefer, Amanda Askell, Anton Bakhtin, Carol Chen, Zac Hatfield-Dodds, Danny Hernandez, Nicholas Joseph, et al. 2024. Towards measuring the representation of subjective global opinions in language models. In *First Conference on Language Modeling*.

Jatin Gupta, Akhil Sharma, Saransh Singhania, Mohammad Adnan, Sakshi Deo, Ali Imam Abidi, and Keshav Gupta. 2025. Large language models acing chartered accountancy.

Tahir Javed, Janki Nawale, Eldho George, Sakshi Joshi, Kaushal Bhogale, Deovrat Mehendale, Ishvinder Sethi, Aparna Ananthanarayanan, Hafsah Faquih, Pratiti Palit, et al. 2024. Indicvoices: Towards building an inclusive multilingual speech dataset for indian languages. In *Findings of the Association for Computational Linguistics ACL 2024*, pages 10740–10782.

Rebecca L Johnson, Giada Pistilli, Natalia Menédez-González, Leslye Denisse Dias Duran, Enrico Panai, Julija Kalpokiene, and Donald Jay Bertulfo. 2022. The ghost in the machine has an american accent: value conflict in gpt-3.

Abhinav Joshi, Shounak Paul, Akshat Sharma, Pawan Goyal, Saptarshi Ghosh, and Ashutosh Modi. 2024. Il-tur: Benchmark for indian legal text understanding and reasoning. In *Proceedings of the 62nd Annual Meeting of the Association for Computational Linguistics (Volume 1: Long Papers)*, pages 11460–11499.

Sankalp KJ, Ashutosh Kumar, Laxmaan Balaji, Nikunj Kotecha, Vinija Jain, Aman Chadha, and Sreyoshi Bhaduri. 2025. Indicmmlu-pro: Benchmarking indic large language models on multi-task language understanding. *CoRR*, abs/2501.15747.

Team Krutrim. 2025. Bharatbench: Comprehensive multilingual multimodal evaluations of foundation ai models for indian languages.

Haonan Li, Yixuan Zhang, Fajri Koto, Yifei Yang, Hai Zhao, Yeyun Gong, Nan Duan, and Timothy Baldwin. 2024. Cmmlu: Measuring massive multitask language understanding in chinese.

Percy Liang, Rishi Bommasani, Tony Lee, Dimitris Tsipras, Dilara Soylu, Michihiro Yasunaga, Yian Zhang, Deepak Narayanan, Yuhuai Wu, Ananya Kumar, et al. 2023. Holistic evaluation of language models. *Trans. Mach. Learn. Res.*

Yang Liu, Meng Xu, Shuo Wang, Liner Yang, Haoyu Wang, Zhenghao Liu, Cunliang Kong, Yun Chen, Maosong Sun, and Erhong Yang. 2024. Omgeval: An open multilingual generative evaluation benchmark for large language models. *arXiv preprint arXiv:2402.13524*.

Ayush Maheshwari, Ashim Gupta, Amrith Krishna, Atul Kumar Singh, Ganesh Ramakrishnan, Anil Kumar Gourishetty, and Jitin Singla. 2024. Samayik: A benchmark and dataset for English-Sanskrit translation.

Ayush Maheshwari, Nikhil Singh, Amrith Krishna, and Ganesh Ramakrishnan. 2022. A benchmark and dataset for post-OCR text correction in Sanskrit.

Arijit Maji, Raghvendra Kumar, Akash Ghosh, Sriparna Saha, et al. 2025. Sanskriti: A comprehensive benchmark for evaluating language models' knowledge of indian culture. *arXiv preprint arXiv:2506.15355*.

RI Masoud, Z Liu, M Ferianc, P Treleaven, and M Rodrigues. 2025. Cultural alignment in large language models: An explanatory analysis based on hofstede's cultural dimensions. In *Proceedings-International Conference on Computational Linguistics, COLING*, pages 8474–8503. Association for Computational Linguistics (ACL).

Kundeshwar Pundalik, Piyush Sawarkar, Nihar Sahoo, Abhishek Shinde, Prateek Chanda, Vedant Goswami, Ajay Nagpal, Atul Singh, Viraj Thakur, Vijay Dewane, Aamod Thakur, Bhargav Patel, Smita Gautam, Bhagwan Panditi, Shyam Pawar, Madhav Kotcha, Suraj Racha, Saral Sureka, Pankaj Singh,

Rishi Bal, Rohit Saluja, and Ganesh Ramakrishnan. 2025. Param-1 BharatGen 2.9b model.

SarvamAI. Sarvam AI | Sovereign Indian AI Ecosystem for LLMs, Agents, and AI Assistants — sarvam.ai. https://www.sarvam.ai/. [Accessed 20-08-2025].

Abhishek Kumar Singh, Vishwajeet Kumar, Rudra Murthy, Jaydeep Sen, Ashish Mittal, and Ganesh Ramakrishnan. 2025. Indic qa benchmark: A multilingual benchmark to evaluate question answering capability of llms for indic languages. In *Findings of the Association for Computational Linguistics: NAACL 2025*, pages 2607–2626.

Harman Singh, Nitish Gupta, Shikhar Bharadwaj, Dinesh Tewari, and Partha Talukdar. 2024. Indicgenbench: A multilingual benchmark to evaluate generation capabilities of llms on indic languages. In *Proceedings of the 62nd Annual Meeting of the Association for Computational Linguistics (Volume 1: Long Papers)*, pages 11047–11073.

Guijin Son, Hanwool Lee, Sungdong Kim, Seungone Kim, Niklas Muennighoff, Taekyoon Choi, Cheonbok Park, Kang Min Yoo, and Stella Biderman. 2025. Kmmlu: Measuring massive multitask language understanding in korean. In *Proceedings of the 2025 Conference of the Nations of the Americas Chapter of the Association for Computational Linguistics: Human Language Technologies (Volume 1: Long Papers)*, pages 4076–4104.

Aarohi Srivastava, Abhinav Rastogi, Abhishek Rao, Abu Awal Md Shoeb, Abubakar Abid, Adam Fisch, Adam R Brown, Adam Santoro, Aditya Gupta, Adrià Garriga-Alonso, et al. 2023. Beyond the imitation game: quantifying and extrapolating the capabilities of language models. *Transactions on Machine Learning Research*, 2023(5):1–95.

Llama 3 Team. 2024. The llama 3 herd of models.

Team-Gemma2. 2024. Gemma 2: Improving open language models at a practical size.

Team-Qwen3. 2025. Qwen3 technical report.

Sshubam Verma, Mohammed Safi Ur Rahman, Vishwajeet Kumar, Rudra Murthy Venkataramana, and Jaydeep Sen. 2025. Milu: A multi-task indic language understanding benchmark. In *Annual Conference of the North American Chapter of the Association for Computational Linguistics*.

Ishaan Watts, Varun Gumma, Aditya Yadavalli, Vivek Seshadri, Manohar Swaminathan, and Sunayana Sitaram. 2024. Pariksha: A large-scale investigation of human-llm evaluator agreement on multilingual and multi-cultural data. In *Proceedings of the 2024 Conference on Empirical Methods in Natural Language Processing*, pages 7900–7932.

# Appendix

## 8 Subject-wise Results

In Table 2 and 4, we present subject-wise results of models with parameter size <8B and >8B, respectively.

### 8.1 Models with <8B parameters

Gemma3-4B emerges as the strongest performer on Current Affairs (70.7), Comparative Religions (50.2), and Defence Studies (49.5), indicating that its pre-training and SFT contain a mix of knowledge-rich domains. Qwen3-4B demonstrates competitive performance across multiple subjects, notably achieving 63.9% on Current Affairs and 47.8% on Defence Studies, while Qwen2.5-3B shows more modest results, with Current Affairs at 37.6% and most cultural subjects remaining below 25%.

The India-focused models present mixed results: Sarvam-1 achieves 45.7% on Current Affairs but struggles with specialized domains like Music (24.5%) and Karnatak Music (24.5%), while Param-1 reaches 35.1 on Current Affairs yet underperforms on most humanities subjects. Cultural and artistic categories remain challenging across all small models: Music scores range from 21.5 to 28.2, Karnatak Music from 27.6 to 31.4, and Percussion Instruments from 25 to 34.9, with Gemma3-4B consistently leading these specialized domains despite absolute accuracies remaining below 35.

The pattern reinforces that even within the sub-8B tier, pretraining mixture quality and domain coverage—rather than parameter count alone—determine performance on culturally grounded Indic content, with Gemma3-4B's superior showing suggesting more balanced exposure to academic and cultural material during training.

### 8.2 Models with >8B parameters

Performance across models with parameter sizes greater than 8B demonstrates clearer scaling benefits on contemporary and knowledge-intensive subjects, though cultural domains remain challenging. Llama-3.3-70B achieves the strongest results on Current Affairs (90.2), Defence Studies (67.4), Yoga (62.8), and Economics (54.6), indicating that its 15T-token pretraining budget and scale provide advantages on well-represented factual domains. Gemma-3-27B leads across multiple humanities subjects despite fewer parameters, achieving top scores in Comparative Religions (70.3), Education (65.1), Sociology (63.1), Anthropology (60), Psychology (57.5), Political Science (55.2), History (52), and Indian Culture (51.9), suggesting that its 14T-token training mixture and SFT training was particularly well-optimized for academic content.

Mistral-3.1-24B demonstrates competitive performance on several specialized topics, notably leading in Music (36.3), Yoga (57.4), and Defence Studies (58.3), while Sarvam-M shows strength in Current Affairs (88.2) and Defence Studies (59.1), reflecting its post-training focus on Indian content. The MoE model Qwen3-30B-A3B maintains solid performance across most categories with only 3B active parameters, achieving 66.1 on Current Affairs and 64.5 on Comparative Religions, demonstrating efficient knowledge capture through expert routing. However, music-related subjects remain uniformly difficult: Music scores range from 25.4 to 36.3, Percussion Instruments from 29.7 to 35.9, and Karnatak Music from 28.7 to 36, with even the largest models struggling to exceed 36 accuracy on these culturally specific domains.

The results confirm that while scale helps on mainstream knowledge tasks, performance on specialized Indic cultural content depends more critically on targeted domain representation in pretraining mixtures, with Gemma-3-27B's humanities strength and persistent music-domain weaknesses across all models.

## 9 Zero Shot Prompt

In Table 5, we provide the zero-shot prompt used while evaluating all the models.

| Subject | Gemma3-1b | Sarvam-1 | Llama-3.2-1B | Param-1 | Qwen2.5-3B | Llama-3.2-3B | Gemma3-4b | Qwen3-4B |
|---|---|---|---|---|---|---|---|---|
| Current Affairs | 28.5 | 45.7 | 38.2 | 35.1 | 37.6 | 56.8 | **70.7** | 63.9 |
| Comparative Religions | 23.3 | 40.7 | 31.9 | 32.5 | 24.9 | 35.6 | **50.2** | 44.8 |
| Defence Studies | 28 | 42.2 | 33.6 | 31.9 | 25 | 36.5 | **49.5** | 47.8 |
| Education | 20.7 | 34 | 29 | 31 | 21.3 | 33.6 | **47.2** | 43.8 |
| Sociology | 23.8 | 32.7 | 30.6 | 33.3 | 20.7 | 29.6 | **42.7** | 39.2 |
| Yoga | 23 | 32.3 | 33.8 | 23.6 | 16.6 | 27.8 | 41.7 | **44.1** |
| Anthropology | 26.8 | 31.5 | 29.5 | 29.1 | 23.7 | 28.7 | 40.3 | **42.2** |
| Tribal Language | 25.2 | 32.6 | 32.6 | 32.2 | 23.4 | 35 | **41.9** | 38.3 |
| Psychology | 24.9 | 28.8 | 28.4 | 26.9 | 20.9 | 29 | 37.7 | **41.3** |
| Economics | 27.1 | 32 | 30.6 | 24.7 | 24 | 28.3 | 39.4 | **40** |
| Political Science | 22.7 | 32.8 | 24.9 | 26.4 | 17.4 | 25.7 | **37.9** | 34.2 |
| Philosophy | 18 | 28.6 | 27.7 | 24.2 | 19.5 | 26.7 | 35.6 | **37** |
| History | 22.6 | 28.8 | 28.1 | 27 | 21.1 | 26.8 | **34.9** | 34 |
| Indian Culture | 23.8 | 27.7 | 23.9 | 26 | 21.4 | 27.9 | **33.9** | 32 |
| Archaeology | 25.8 | 29 | 29.4 | 27.5 | 23.4 | 28.8 | **34.2** | 33.2 |
| Drama & theatre | 24.2 | 30.2 | 23.6 | 21.1 | 22.7 | 26.2 | 35.6 | **36.1** |
| Law | 22.2 | 31.3 | 29.1 | 26.8 | 20.7 | 30.3 | 36.2 | **36.8** |
| Rabindra Sangeet | 26 | 26.9 | 28 | 29.6 | 26.4 | 28.2 | 31.3 | **32.6** |
| Music | 21.9 | 24.5 | 21.7 | 25.9 | 21.5 | 24 | 28.2 | **28.2** |
| Percussion Instruments | 28.9 | 26.3 | 30.5 | 27.9 | 28 | 25.5 | **34.9** | 33.6 |
| Karnatak Music | 27.6 | 24.5 | 26.6 | 29.3 | 27.9 | 29.5 | **31.4** | 30.8 |

Table 4: We report the accuracy across different subjects for LLMs with parameter size <8B. The best-performing model for each subject is indicated in **bold**.

**Zero-Shot Prompting**

```
Prompt = f"""Question: {['question_text']}

Options:
A) {[‘option_a’]}
B) {[‘option_b’]}
C) {[‘option_c’]}
D) {[‘option_d’]}

Given the above question and multiple options, select the correct answer. Keep your response only in English with one of the letters corresponding to the options A, B, C, or D. Do not write anything else.”””.
```

Table 5: Zero-Shot prompt applied across all models for evaluation

## 10 Question Types

In Table 6 and 7, we present question type-wise results for all the evaluated models with parameter size <8B and >8B, respectively.

Among smaller models (<8B parameters), Qwen3-4B demonstrates the strongest overall performance, leading in Assertion & Reason (49.7), Identify Incorrect Statement (43.9), Fill in the Blank (45.7), Normal MCQ (41.9), and Ordering (36.6), suggesting that its instruction-tuning pipeline effectively covered diverse reasoning formats. Gemma3-4B shows competitive results on Assertion & Reason (46.5) and Fill in the Blank (52.2), while Sarvam-1 achieves respectable scores on Assertion & Reason (40.5) and Normal MCQ (33), though both trail Qwen3-4B across most categories.

In the larger model tier (>8B parameters), Llama-3.3-70B dominates Assertion & Reason (61.8) and Normal MCQ (57.8), while Gemma-3-27B excels in Identify Incorrect Statement (60.4), Fill in the Blank (58.7), Match the List (52.7), and Ordering (50.1), demonstrating format-specific strengths that reflect different SFT priorities. The substantial gap between Gemma-3-27B (60.4) and Llama-3.3-70B (45.7) on Identify Incorrect Statement tasks is particularly notable, suggesting that error-detection capabilities depend more on instruction-tuning composition than raw scale. Across all models, ordering tasks remain the most challenging format, with even top performers struggling to exceed 50 - Gemma-3-27B reaches 50.1 and Qwen3-4B achieves 36.6 - indicating that temporal and procedural reasoning on Indic content

| Question Type | Gemma3-1b | Sarvam-1 | Llama3.2-1B | Param-1 | Qwen2.5-3B | Llama3.2-3B | Gemma3-4b | Qwen3-4B |
|---|---|---|---|---|---|---|---|---|
| Assertion & Reason | 39.1 | 40.5 | 34.4 | 28.2 | 20.6 | 37.6 | 46.5 | **49.7** |
| Incorrect Statement | 18.3 | 31.4 | 23 | 21.7 | 16.7 | 24.7 | **42.9** | 38 |
| Fill in the Blank | 19.6 | 39.1 | 19.6 | 21.3 | 15.2 | 23.9 | **52.2** | 45.7 |
| Normal MCQ | 23.1 | 33 | 30.2 | 29.9 | 24.3 | 31.1 | **41.9** | 38.3 |
| Match the List | 22.2 | 22.9 | 25 | 27.1 | 23.6 | 28.6 | 28.2 | **35.5** |
| Ordering | 24.1 | 22.8 | 27.1 | 27.2 | 24 | 26.6 | 32.9 | **36.6** |

Table 6: We present the average accuracy across all subjects for various question types, focusing on the models with parameter size <8B. The best-performing model for each question type is indicated in **bold**

| Question Type | Aya-8b | Llama-8B | Mistral-24B | Sarvam-m | Gemma-27b | Qwen-MoE | Aya-32B | Llama-70B |
|---|---|---|---|---|---|---|---|---|
| Assertion & Reason | 40.6 | 39.2 | 44.9 | 51.6 | 60.5 | 47.2 | 54.4 | **61.8** |
| Incorrect Statement | 29.1 | 29.6 | 52.4 | 44.1 | **60.4** | 53.6 | 48.8 | 45.7 |
| Fill in the Blank | 37 | 39.1 | 50 | 45.7 | **58.7** | 54.3 | 43.5 | 54.3 |
| Normal MCQ | 37.7 | 41 | 51.9 | 49.9 | 56.6 | 48.7 | 50.1 | **57.8** |
| Match the List | 25.8 | 26.2 | 50.7 | 48.8 | **52.7** | 45.5 | 29.3 | 41.9 |
| Ordering | 31.2 | 31.8 | 51 | 46.7 | **50.1** | 47.9 | 33.5 | 49 |

Table 7: We present the average accuracy across all subjects for various question types, focusing on the models with parameter size >8B. The best-performing model for each question type is indicated in **bold**

remains underrepresented in current instruction datasets. The pattern confirms that question-type performance is shaped primarily by SFT design choices rather than parameter count alone, with Qwen3-4B's consistent leadership among small models and Gemma-3-27B's dominance in structured reasoning tasks highlighting the importance of targeted instruction-tuning for diverse format coverage.

## 11 Examples questions in ParamBench

Table 8 presents examples of six distinct types of questions used in our benchmark dataset. For each type, two representative questions have been selected to illustrate the structure, content, and answer format.

| Type | Question | Options | Ans. |
| --- | --- | --- | --- |
| MCQ | पाषणकालीन उपकरणों की उपयोगिता की अध्ययन विधि है : | (a) स्तर विज्ञान<br>(b) सूक्ष्म चिह्नीय अध्ययन<br>(c) शुल्कन प्रयोग<br>(d) प्रारूपकीय विज्ञान | (b) |
| MCQ | निम्नांकित मृदभांड परम्पराओं में से कौन महाभारत काल से जुड़ा हुआ है ? | (a) उत्तरी काले चमकीले मृदभांड<br>(b) कृष्ण–लोहित मृदभांड<br>(c) गैरिक मृदभांड<br>(d) चित्रित धूसर मृदभांड | (d) |
| Incorrect Statement Identification | उस कूट को चिन्हित करें जिसमें सही अभिकथन न हो : | (a) एक संयोजक सत्य है यदि इसके सभी संघटक सत्य हैं अन्यथा यह असत्य है।<br>(b) प्रत्येक यौगिक अभिकथन एक सत्यता–फलन अभिकथन होता है।<br>(c) द्विमूल्याश्रित तर्कशास्त्र में प्रत्येक अभिकथन या तो सत्य होता हैं या असत्य।<br>(d) एक सरल अभिकथन वह अभिकथन है जिसका संघटक इसके भाग के रूप में अन्य कोई अभिकथन नहीं होता। | (b) |
| Incorrect Statement Identification | निम्नलिखित में से कौन सा युग्म सही सुमेलित नहीं है ? | (a) खारवेल का हाथीगुम्फा – पार्श्वनाथ अभिलेख<br>(b) चन्द्रगुप्त द्वितीय का – वीरसेन उदयगिरि गुफा अभिलेख साब<br>(c) रुद्रदामन का – तुषास्फ जूनागढ़ शिलालेख<br>(d) कुमारगुप्त एवं – वत्सभट्टि बन्धुवर्मा का मन्दसौर प्रस्तर अभिलेख | (a) |
| List-based Matching | सूची–I को सूची–II के साथ सुमेलित करें।<br>सूची–I: (a) पक्षधर्मता, (b) विपक्षसत्त्व, (c) बाधित, (d) विरुद्ध<br>सूची–II: (i) अग्नि शीतल है, (ii) शब्द शाश्वत है क्योंकि यह उत्पन्न होता है, (iii) पर्वत पर धूम्र है, (iv) जलाशय में अग्नि है | (a) (iv) (i) (iii) (ii)<br>(b) (i) (ii) (iii) (iv)<br>(c) (ii) (iii) (i) (iv)<br>(d) (iii) (iv) (i) (ii) | (d) |
| List-based Matching | सूची–I और सूची–II को सुमेलित करें।<br>सूची–I: (a) वैयक्तिक प्रत्ययवाद, (b) एकतत्त्व प्रत्ययवाद, (c) आत्मनिष्ठ प्रत्ययवाद, (d) यथार्थवादी प्रत्ययवाद<br>सूची–II: (i) सीमित आत्मा एक का अंश, प्रकार अथवा अभास है। (ii) मूर्तसत्ता वैयक्तिक आत्मत्व है। (iii) वस्तुओं के आदर्श रहित रूपों की यथार्थता को स्थापित करते हैं। (iv) प्रकृति सीमित मन का प्रक्षेपण मात्र है। | (a) (iv) (iii) (ii) (i)<br>(b) (ii) (iv) (i) (iii)<br>(c) (ii) (i) (iv) (iii)<br>(d) (i) (ii) (iii) (iv) | (c) |

| Type | Question | Options | Ans. |
|---|---|---|---|
| Assertion & Reasoning | दिये गये अभिकथन (A) और तर्क (R) की परीक्षा आगमनात्मक अनुमान के आलोक में करें और नीचे दिये गये कूट में से सही का चयन करें। अभिकथन (A) : आगमनात्मक अनुमान में ज्ञात से अज्ञात की ओर जाते हैं। तर्क (R) : आगमनात्मक अनुमान में किसी जाति विशेष के सभी सदस्यों के निर्णय के द्वारा उस जाति विशेष के सभी सदस्यों के बारे में निर्णय तक पहुँचते हैं। कूट : | (a) दोनों (A) और (R) सही व्याख्या है<br>(b) दोनों (A) और (R) सही व्याख्या नहीं है<br>(c) (A) सही, (R) गलत<br>(d) (R) सही, (A) गलत | (c) |
| Assertion & Reasoning | नीचे एक अभिकथन (A) और एक कारण (R) दिये गये हैं। उन पर विचार कीजिये और नीचे दिये गये कूट से सही विकल्प का चयन कीजिये। अभिकथन (A): परमाणु की सत्ता अवश्य स्वीकार की जानी चाहिये। तर्क (R): द्वयणुक सावयव है। कूट : | (a) (A) और (R) दोनों सही हैं और (R), (A) का सही आधार है।<br>(b) (A) और (R) दोनों सही हैं और (R), (A) का सही आधार नहीं है।<br>(c) (A) सही है और (R) गलत है।<br>(d) (A) गलत है और (R) सही | (a) |
| Ordering | कालसमयानुसार ग्रन्थों का सही क्रम चुनिए : | (a) संगीत मकरंद, राग–विबोध, नारदीय–शिक्षा, राग–तरंगिनी<br>(b) नारदीय–शिक्षा, संगीत मकरंद, राग–विबोध, राग–तरंगिनी<br>(c) राग–तरंगिनी, संगीत मकरंद, नारदीय–शिक्षा, राग–विबोध<br>(d) नारदीय–शिक्षा, संगीत मकरंद, राग–तरंगिनी, राग–विबोध | (d) |
| Ordering | कालक्रमानुसार सही क्रम चुनिए : | (a) खुदा बक्श, फैयाज़ खान, गुलाम अब्बास, शराफत हुसैन<br>(b) खुदा बक्श, गुलाम अब्बास, फैयाज़ खान, शराफत हुसैन<br>(c) फैयाज़ खान, शराफत हुसैन, खुदा बक्श, गुलाम अब्बास<br>(d) शराफत हुसैन, खुदा बक्श, फैयाज़ खान, गुलाम अब्बास | (b) |
| Fill in the blanks | सखी–कुंधे नट ________ की रंगमंचीय प्रस्तुति है। | (a) असम<br>(b) आंध्र प्रदेश<br>(c) पंजाब<br>(d) ओडिशा | (d) |

| Type | Question | Options | Ans. |
| --- | --- | --- | --- |
| Fill in the blanks | निम्नलिखित में रिक्त स्थान की पूर्ति करें: अन्तःप्रज्ञावाद, अन्तःप्रज्ञात्मता ________ | (a) के समकक्ष<br>(b) के समान<br>(c) से भिन्न<br>(d) इनमें से कोई नहीं | (c) |